\pdfoutput=1

\documentclass[11pt]{article}

\usepackage{acl}
\usepackage{algorithm}
\usepackage{algorithmic}
\usepackage{times}
\usepackage{latexsym}
\usepackage{enumitem}
\usepackage[T1]{fontenc}
\usepackage{longtable}

\usepackage[utf8]{inputenc}
\usepackage{multirow}
\usepackage{microtype}
\usepackage{graphicx}
\usepackage{inconsolata}
\usepackage{pgfplots}
\usepackage{amssymb}
\usepackage{subfigure}
\usepgfplotslibrary{groupplots}

\title{Can LLMs Learn from Previous Mistakes? Investigating LLMs' Errors to Boost for Reasoning}

\author{
  Yongqi Tong\textsuperscript{1}, 
  Dawei Li\textsuperscript{1},
  Sizhe Wang\textsuperscript{2},
  Yujia Wang\textsuperscript{1}, 
  Fei Teng\textsuperscript{1}, 
  Jingbo Shang\textsuperscript{1}\thanks{\textdagger Corresponding author.}\\
  \textsuperscript{1}University of California, San Diego, \texttt{\{yotong, dal034, yuw103, feteng, jshang\}@ucsd.edu} \\
  \textsuperscript{3}University of Southern California, \texttt{sizhewan@usc.edu} \\
}

\begin{document}
\maketitle
\begin{abstract}

Large language models~(LLMs) have demonstrated striking reasoning capability.
Recent works have shown the benefits to LLMs from fine-tuning golden-standard Chain-of-Thought (CoT) rationales or using them as correct examples in few-shot prompting. 
While humans can indeed imitate correct examples, 
learning from our mistakes is another vital aspect of human cognition. 
Hence, a question naturally arises: \textit{can LLMs learn and benefit from their mistakes, especially for their reasoning? }
This study investigates this problem from both the prompting and model-tuning perspectives.
We begin by introducing \textsc{CoTErrorSet}, a new benchmark with 558,960 questions, each designed with both correct and error references, and demonstrating the types and reasons for making such mistakes.
To explore the effectiveness of those mistakes, we design two methods: 
(1) \textbf{Self-rethinking} prompting guides LLMs to rethink whether they have made similar previous mistakes;
and (2) \textbf{Mistake tuning} involves finetuning models in both correct and incorrect reasoning domains, rather than only tuning models to learn ground truth in traditional methodology.
We conduct a series of experiments to prove LLMs can obtain benefits from mistakes in both directions. 
Our two methods offer potentially cost-effective strategies by leveraging errors to enhance reasoning capabilities, which costs significantly less than creating meticulously hand-crafted golden references. 
We ultimately make a thorough analysis of the reasons behind LLMs' errors, which provides directions that future research needs to
overcome.
\textsc{CoTErrorSet} will be published soon on \texttt{\url{https://github.com/YookiTong/Learn-from-Mistakes-CotErrorSet}}. 

\end{abstract}

\section{Introduction}

\begin{figure}[t]
    \centering
    \scalebox{0.45}{\includegraphics{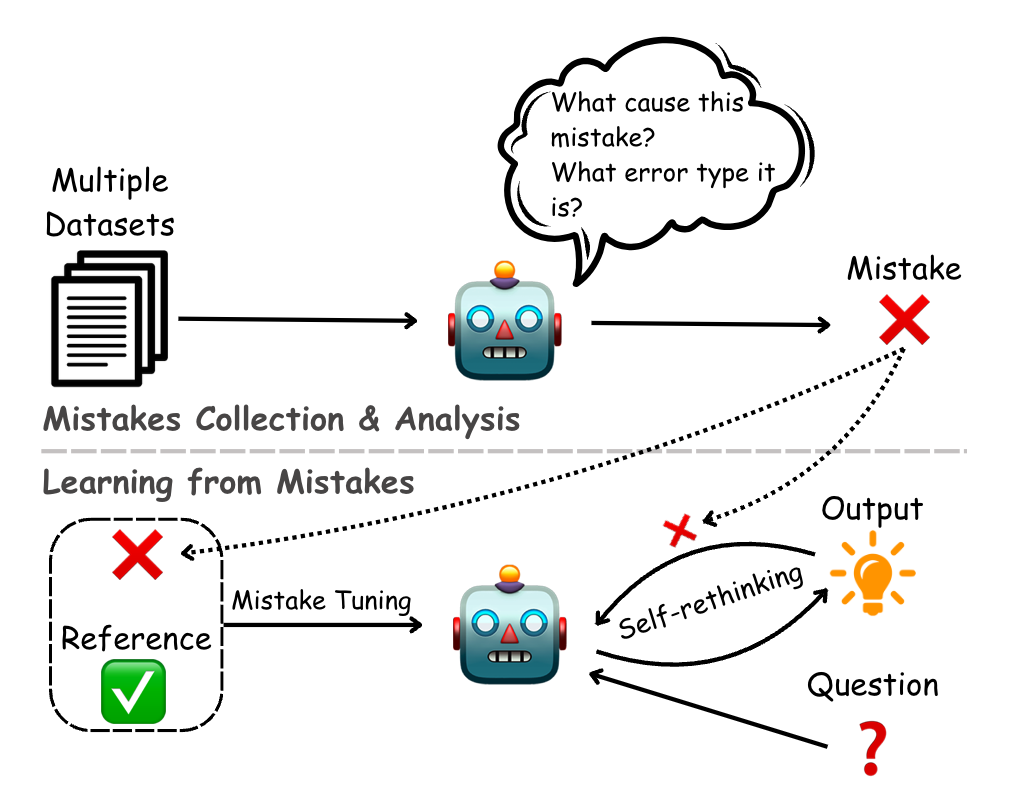}}
    \caption{The overview pipeline of our work includes (1). Mistake collection and analysis (Section~\ref{A Novel Dataset}). (2) Two novel methods to instruct LLMs to learn from mistakes(Section~\ref{Our Methodology: Self-rethinking} and Section~\ref{Our Methodology: Mistake Tuning}).}
    \label{fig:enter-label}
\end{figure}

Large language models~(LLMs)~\citep{brown2020language,zhang2022opt,anil2023palm,touvron2023llama} have demonstrated strong capabilities across various tasks and applications~\citep{liang2022holistic,chang2023survey}.
To further unleash the reasoning abilities of LLMs and align their thinking process with humans, many recent studies explored Chain-of-Thought (CoT)-based prompting~\citep{wei2022chain,wang2022self,li2023dail,tong2023eliminating,yao2023tree,besta2023graph} to instruct LLMs to solve the given problem with human-like logic.
Besides logical step-by-step thinking, another critical learning pattern of us humans is to rethink and learn from our previous mistakes so that avoid repeating the same mistakes in the future~\citep{mercer2008talk, reich2023overcome}.
However, few studies have focused on systematically understanding what kinds of intermediate errors occur in making CoT procedures and whether LLMs can learn from those mistakes.
To address these issues, we aim to explore the potential of LLMs to effectively utilize their previous mistakes to boost reasoning.

To enhance the scalability and efficiency of analyzing and learning from the mistakes of LLMs, we began by collecting a vast dataset of LLMs' reasoning outputs and built~\textsc{CoTErrorSet}, which consists of 609,432 questions sourced from 1060 tasks across diverse domains. 
Each query in this set is meticulously structured, featuring both a manually curated correct reference and the incorrect rationales collected from PaLM2~\citep{anil2023palm}'s responses. 
Furthermore, we prompt the LLMs with the correct reference and the incorrect responses in order to make it reflect why making such mistakes. 
The introspective responses are also collected and subsequently utilized in our work. 
We employ this data for cluster analysis to identify specific details of the errors.

With our \textsc{CoTErrorSet}, we introduce two innovative paradigms, namely \textbf{mistake tuning} and \textbf{self-rethinking}, aimed at efficiently augmenting LLMs by leveraging their historical errors during both tuning and inference stages.
Diverging from the conventional approach of only relying on correct rationales in traditional supervised fine-tuning, our mistake tuning strategy incorporates combinations of both correct references and incorrect rationales.
To facilitate the learning process for LLMs, we introduce the prefixes \textit{[CORRECT RATIONALE]} and \textit{[INCORRECT RATIONALE]} before the corresponding rationales.
Intuitively, this prompt tuning facilitates LLMs to distinguish between correct and incorrect rationales while avoiding corruption from the incorrect ones with the two separated prefixes.
For self-rethinking, inspired by contrastive in-context learning~\citep{gao2024customizing}, we expose LLMs to both correct and incorrect rationales in demonstration samples.
After obtaining the initial answer output by the LLM, we iteratively prompt it to rethink and rectify the result based on the historical mistakes.
To manage computational resources and prevent potential loops, we implement a threshold, limiting the number of times the model can engage in self-rethinking and corrections.
Figure~\ref{fig:enter-label} gives an overview pipeline of our work.

To substantiate the efficacy of our proposed methodologies and to delve into the learning capabilities of LLMs from their mistakes, we undertake experiments encompassing diverse reasoning tasks and LLMs of varying sizes.
The application of our methods consistently yields performance enhancements across a spectrum of tasks, underscoring the effectiveness and broad applicability of our approaches in leveraging LLMs' mistakes during both the tuning and inference stages.
Additionally, we conduct thorough analyses of the error types exhibited by LLMs, offering comprehensive insights and guidance on mitigating the most prevalent errors in these models.

In general, our contributions are as follows:

\begin{itemize}
    \item A large-scale error set, \textsc{CoTErrorSet}, is constructed for scalable analysis and learning from the LLMs' mistakes.
    \item We novelly designed two paradigms for LLMs to utilize and learn from their previous mistakes at both fine-tuning and inference stages.
    \item With extensive experiments, we validate the effectiveness of our proposed methods and provide further hints based on analysis of LLMs' error types.
\end{itemize}

\section{Related Work}

\textbf{Human-like Reasoning with LLMs.} CoT~\cite{wei2022chain} demonstrate the great potential of equipping LLMs with human-like reasoning capability.
Following them, various logical and structural reasoning strategies~\cite{wang2022self,zhou2022least,creswell2022faithful,besta2023graph,li2023making,lightman2023let} are proposed to align LLMs' thinking processes with humans.
These enhanced reasoning approaches have been adopted in different tasks and areas, including commonsense reasoning~\cite{geva2021did,ahn2022can}, logical reasoning~\cite{pan2023automatically,lei2023boosting} and mathematical reasoning~\cite{cobbe2021training,hendrycks2021measuring} and achieved promising performance.
In this work, we aim to investigate whether LLMs can benefit from rethinking and learning from previous mistakes, which is one of the most important learning patterns of humans.


\textbf{Refined Reasoning Errors. } 
Several studies focus on adjusting their reasoning pathways to arrive at better solutions. 
\citet{huang2022large} introduce self-improve that employs CoT plus self-consistency to obtain high-confidence solutions on a large set of unlabeled questions.  
The self-generated content is then used for fine-tuning in subsequent iterations, thereby further augmenting its reasoning capabilities. 
\citet{madaan2023self} propose a self-refine technique that encourages LLMs to autonomously correct their outputs without the need for external data or feedback. 
However, it has been argued by some researchers that LLMs face challenges in self-correcting their responses in the absence of external feedback, and under certain conditions, such attempts might even deteriorate their performance~\citep{huang2023large}. 
Based on that, \citet{an2023learning} suggest fine-tuning LLMs using pairs consisting of errors and their respective corrections generated by GPT-4 as a supervisory mechanism. 
Nevertheless, our work is pioneering in highlighting the impact of exposing mistake examples on in-context learning. 
Furthermore, our experiments reveal that in the process of model tuning, learning from mistakes can inherently enhance itself by merely being exposed to correct examples and errors, without depending on explicit corrections from teacher models.

\section{A Novel Dataset: \textsc{CoTErrorSet}}
\label{A Novel Dataset}

\subsection{Dataset Construction}
In order to investigate whether incorrect rationales can also contribute to LLMs' reasoning performance, we introduce \textsc{CoTErrorSet}, a novel benchmark based on the source of \textsc{CoT-Collection}~\citep{kim2023cot}, built upon various domains, including multiple-choice QA, extractive QA, closed-book QA, formal logic, natural language inference, and arithmetic reasoning. 
Our dataset's question and reference are obtained from the following datasets: QASC~\citep{khot2020qasc}, AQuA~\citep{ling2017program}, GSM8K~\citep{cobbe2021training}, QED~\citep{lamm2021qed}, StrategyQA~\citep{geva2021did}, SenseMaking~\citep{wang2019does}, CREAK~\citep{onoe2021creak}, e-SNLI~\citep{camburu2018snli} and ECQA~\citep{aggarwal2021explanations}. 
Each task within this collection is systematically organized to include a question and a correct reference, followed by an incorrect response and the demonstrations why making such mistakes. 
The errors and demonstrations are both generated from PaLM2.

\begin{figure}[t]
    \centering
    \scalebox{0.4}{\includegraphics{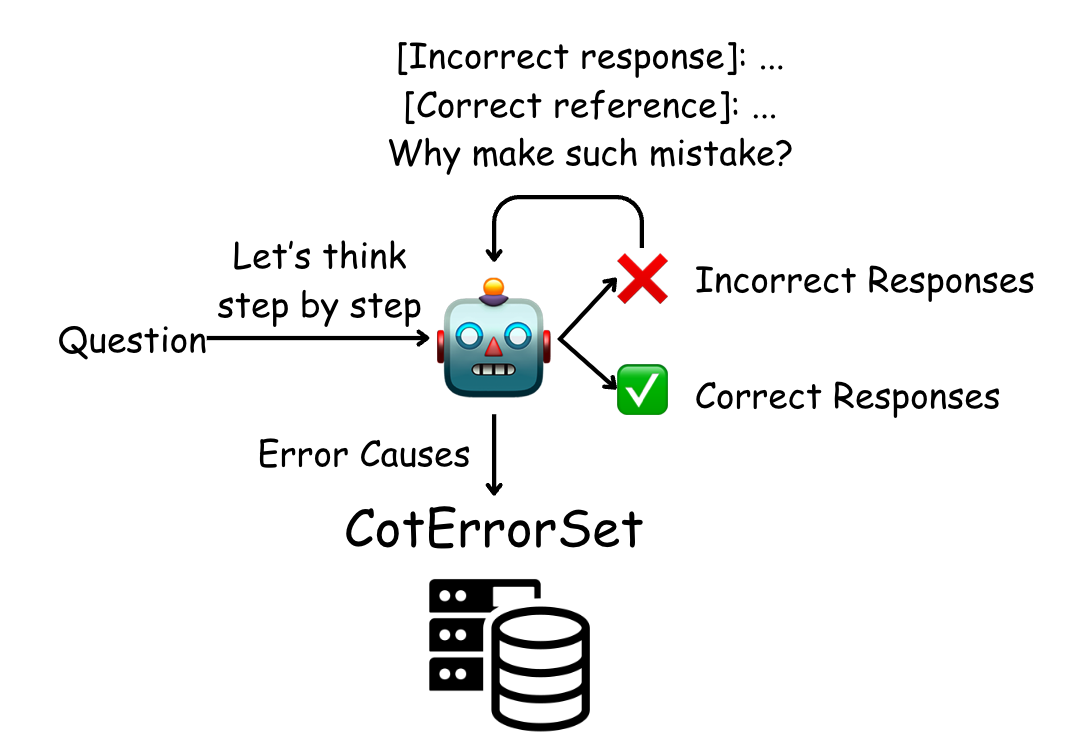}}
    \caption{The pipeline to construct \textsc{CoTErrorSet}. By providing PaLM2 with the correct reference and the incorrect response generated by itself, we prompt it to introspect and grasp the underlying reasons for its errors.}
    \label{fig:prompt_construction}
\end{figure}

\textsc{CoTErrorSet} diverges from traditional CoT datasets by employing PaLM2's mistakes and the reasons behind them.
We utilized PaLM2 to generate rationales for each question in the dataset, focusing specifically on collecting incorrect rationales.
Recent research has demonstrated LLMs' capability to provide high-quality data~\cite{li2024contextualization,tong2024optimizing,li2024dalk} and feedback~\cite{pan2024automatically,tan2024large}.
Following this idea, we provide PaLM2 with both correct references and its incorrect answers to demonstrate and reflect why it makes such mistakes. 
The steps of the construction process are shown in Figure~\ref{fig:prompt_construction}.
This systematic collection of incorrect rationales can make \textsc{CoTErrorSet} a promising benchmark in providing future improvements from a different perspective. 
One example is shown in Table~\ref{table:mistake}. 

\begin{table}[h]
\centering
\small
\begin{tabular}{p{0.9\columnwidth}}
\hline
\textbf{Questions: } Combine facts and answer this: Which meridian extends across Europe, the Mediterranean Sea, Africa, Asia, the Pacific Ocean, North America, and the Atlantic Ocean? \\
\hline
\textbf{Target:} The Cimarron meridian \\\hline
\textbf{Reference:} The Cimarron meridian extends across Europe, the Mediterranean Sea, Africa, Asia, the Pacific Ocean, North America and the Atlantic Ocean.\\\hline
\textbf{Incorrect Rationale:} The 180th meridian extends across Europe, the Mediterranean Sea, Africa, Asia, the Pacific Ocean, North America and the Atlantic Ocean.\\\hline
\textbf{Error Causes:} Making mistakes in incorrect rationales, such as claiming the 180th meridian extends across various continents and oceans, can lead to significant misinformation and confusion. This particular error demonstrates a fundamental misunderstanding of geography, as the 180th meridian primarily runs through the Pacific Ocean and does not cross the regions listed. Such inaccuracies underscore the importance of fact-checking in educational content to prevent the spread of misconceptions. Correcting these mistakes not only clarifies the factual information but also serves as a valuable learning opportunity, emphasizing the need for accuracy and critical evaluation of information.\\\hline
\end{tabular}
\caption{An example in \textsc{CoTErrorSet}. The content of \textit{Incorrect Rationale} and \textit{Error Causes} are generated by PaLM2 as indicated in Figure~\ref{fig:prompt_construction}. }
\label{table:mistake}
\end{table}

\subsection{Error Analysis with \textsc{CoTErrorSet}}

\begin{figure}[h]
    \centering
    \scalebox{0.3}{
    \includegraphics{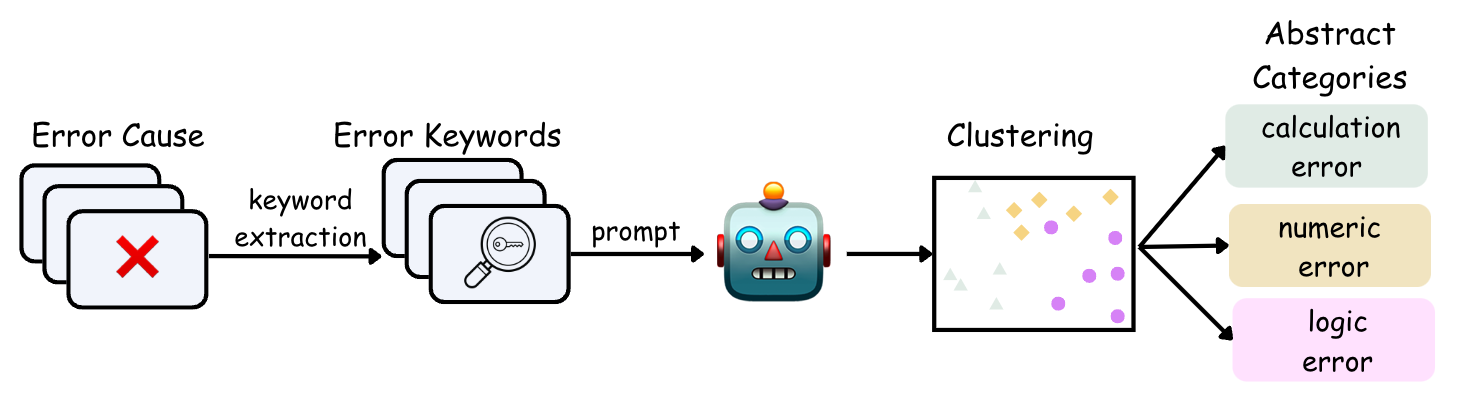}
    }
    \caption{Our pipeline for clustering PaLM2's mistakes. }

    \label{fig:clustering}
\end{figure}

After collecting the \textsc{CoTErrorSet} dataset, we observe that the error types in it are very intricate and diverse.
The intricacy poses obstacles to subsequent enhancement efforts.
In order to tackle this issue and gain a more overarching understanding of LLMs' error types, we utilize an LLM-based unsupervised clustering approach shown in Figure~\ref{fig:clustering} to match diverse error types into more general categories.

\begin{table*}[h]
    \centering
    \small
    \noindent\rule{\textwidth}{1pt}
    \begin{itemize}[leftmargin=*,label={},topsep=0pt, partopsep=0pt, parsep=0pt, itemsep=0pt]
    \item \textbf{Step 1: CoT reasoning} \\
        \begin{itemize} [topsep=0pt, partopsep=0pt, parsep=0pt, itemsep=0pt]
            \item \textit{Question:} Natalia sold clips to 48 of her friends in April, and then she sold half as many clips in May. How many clips did Natalia sell altogether in April and May?
            \item \textit{Answer: } Let's think step by step.
            \item \textcolor{red}{Response: Natalia sold 48 * 2 = 96 clips in May. Natalia sold 48+96 = 144 clips altogether in April and May. }
        \end{itemize}
    \noindent\rule{\textwidth}{1pt}
    
    \item \textbf{Step 2: self-rethinking} \\ 
        \begin{itemize}[topsep=0pt, partopsep=0pt, parsep=0pt, itemsep=0pt]
                \item \textit{Question:} Natalia sold clips to 48 of her friends in April, and then she sold half as many clips in May. How many clips did Natalia sell altogether in April and May?
                \item \textit{Your output: } Natalia sold 48 * 2 = 96 clips in May. Natalia sold 48+96 = 144 clips altogether in April and May. 
                \item \textit{Do you make similar mistakes with the following examples: } \#\#\#Error Type 1: Misapplication of Algebraic Identities: ...
                \item \textcolor{red}{Response: Yes, I make a mistake. }
        \end{itemize}
        
    \noindent\rule{\textwidth}{1pt}
    
    \item \textbf{Step 3: correction (if they think they make mistakes)} \\ 
        \begin{itemize}[topsep=0pt, partopsep=0pt, parsep=0pt, itemsep=0pt]
            \item \textit{So the correct answer is: } 
            \item \textcolor{red}{Response: Natalia sold 48/2 = 24 clips in May. Natalia sold 48+24 = 72 clips altogether in April and May.}
        \end{itemize}
    \end{itemize}
    \noindent\rule{\textwidth}{1pt}
\caption{One example of interactive prompting and responses for self-rethinking. Black texts are the prompting while the red content serves as LLMs' response example.}
\label{tab: dataset_example}
\end{table*}

To be specific, we begin by extracting the specific error keywords from each error cause.
Subsequently, we input all the extracted keywords into the LLMs and prompt them to generate more general categories that encompass the entire spectrum of error names.
Following this automated clustering process, we manually review each cluster, making necessary adjustments to refine the matching results.
Finally, we distill the diverse error types into several abstract categories, such as calculation error, numeric error, and logical error in domains of arithmetic reasoning and logical error, commonsense error, linguistic error, and context error in domains of commonsense reasoning.
A detailed definition of each error category is shown in Appendix~\ref{Definition for Error Categories}.
We put results and analysis in Section~\ref{sec: ana}.






\section{Our Methodology: Self-rethinking}
\label{Our Methodology: Self-rethinking}

Self-rethinking offers an innovative approach to encourage LLMs to consider if they are repeating past errors. 
This method starts with an initial CoT reasoning. Following this, the model uses the provided reasoning outputs and a random selection of examples from \textsc{CoTErrorSet}.  
This step is designed to assess if the model's most recent response includes similar inaccuracies. 
If errors are detected, it will formulate a new rationale and undergo the evaluation process again. This cycle continues until the model deems its latest answer to be correct or it reaches a set limit of evaluation rounds. 
The main goal is to empower the LLM to learn from its errors introspectively and minimize the recurrence of such mistakes.
One example is shown in Table~\ref{tab: dataset_example}. 


The core of self-rethinking lies in the backward-checking stage. In this phase, the LLM reviews its reasoning chain, but with a specific focus on the error types it previously identified. 
This explicit demonstration of errors, coupled with the question, golden reference, and incorrect rationales, is instrumental in enabling the LLM to recognize specific types of mistakes it tends to make. 
This targeted review helps the LLM to not just correct the random errors but to consciously avoid repeating the same types of mistakes it has made in the past.
The process includes a loop for error correction and confirmation. 
If the LLM finds that it has repeated any of the previously identified mistakes, it revisits the reasoning process to correct them. 
Otherwise, the last response is adopted as the final result. 

Moreover, the iterative checking process should have a crucial repeating boundary, denoted as $k$ iterations. 
If the LLM's error-checking and correction cycles surpass this predefined threshold and errors still persist, the process concludes under the assumption that the issue at hand or the error detection might exceed the LLM's current capabilities.
This constraint prevents the LLM from being caught in an endless loop of self-rethinking, ensuring the efficiency and practicality of the reasoning process.


\section{Our Methodology: Mistake Tuning}
\label{Our Methodology: Mistake Tuning}
In order to fully investigate the other potential utilization of our principles, we introduce mistake tuning, which demonstrates our motivation is a broad and pioneering framework not only in the field of in-context learning. 
This approach is designed to finetune LLMs on the combinations of both correct rationales and incorrect mistakes. 
By simply appending prefixes \textit{[CORRECT RATIONALE]} and \textit{[INCORRECT RATIONALE]} before corresponding rationales, mistake tuning can further improve LLMs' abilities to distinguish between correct and incorrect rationale.

Mistake tuning is built upon the foundational motivations and conclusions of self-rethinking, where LLMs can learn from the implicit reasons and types of mistakes they made in CoT reasoning.
This process can be formulated as:

\begin{equation}
    p = [Q \oplus S \oplus R],
\end{equation}

\begin{equation}
    \mathcal{L} = -\sum_{t=1}^{|p|} log P(p_{t}|p_{<t}),
\end{equation}

Where $Q$, $S$ and $R$ represent the given question, special prefix and corresponding rationale respectively. $\oplus$ represents the operation of concatenation.

Mistake tuning presents a cost-effective, straightforward, and efficient alternative. 
Previous work has proven pretraining with some controlled signals based on human feedback can result in LLMs' better ability to generate more satisfactory contents~\citep{korbak2023pretraining, keskar2019ctrl}. 
Hence, incorporating fixed prefixes in finetuning LLMs in the field of reasoning can also help models differentiate information from golden references and mistakes. 
Our results also demonstrate its effectiveness for promoting LLMs' reasoning abilities without additional costs similar to annotating golden reasoning references.

\section{Experiments}

In this section, we conducted a series of experiments to compare the proposed self-rethinking methods with the existing approach on both arithmetic and commonsense reasoning benchmarks. 

\begin{table*}[h]
    \small
    \centering
    \begin{tabular}{c|cccccc}
     \hline  \textbf{Methods} & \textbf{GSM8K} & \textbf{AQuA} & \textbf{MathQA} &  \textbf{OpenbookQA} & \textbf{LogiQA} & \textbf{CR}\\\hline
        Standard Prompting~\citep{brown2020language} & 17.06 & 22.40 & 27.57 & 80.92 & 41.21 & 24.45\\
        CoT~\citep{madaan2023self} & 56.29 & 32.11 & 30.89 & 82.66 & 41.05 & 51.98 \\
        Self-refine~\citep{madaan2023self} & 34.74&	39.92&	\textbf{54.01} &	28.75&	35.99&	12.28\\
        Self-consistency~\citep{wang2022self} & 58.38 & 42.80 & 41.37& 87.61 & 42.88 & 22.58\\
        Self-rethinking (Ours) & \textbf{65.13} & \textbf{44.72} & 43.95 &  \textbf{87.71} & \textbf{49.12} & \textbf{54.53}\\\hline
    \end{tabular}
    \caption{PaLM2's accuracy on the existing baselines and our methods, self-rethinking prompting. Self-rethinking shows consistent improvements but uses less inference time compared with self-consistency. 
    }
    \label{tab: arithmetic reasoning res}
\end{table*}

\begin{table}[h]
    \scalebox{0.85}{
    \small
    \centering
    \begin{tabular}{c|cccc}
        \hline \textbf{Methods} &\textbf{GSM8K} & \textbf{AQuA} &\textbf{MathQA} & \textbf{LogiQA}  \\\hline
        8-shot CoT & 64.56 & 30.65 & 36.21 & 29.57  \\
        8-shot self-rethinking &  \textbf{70.15} & \textbf{34.80} & \textbf{40.56} & \textbf{33.64} \\\hline
    \end{tabular}
        }
    \caption{PaLM2's accuracy results on few-shot Chain-of-Thought(CoT) and our methods, self-rethinking. We select 8-shot examples from the corresponding trainset. Then we collect PaLM2's incorrect rationales of those 8 examples. The part of the original correct reference is CoT's demonstrations. Those generated incorrect rationales serve as demonstrations for the rethink stage. }
    \label{tab:few_shot}

\end{table}

\begin{table}[h]
    \scalebox{0.75}{
        \centering
        \begin{tabular}{c|cccc}
            \hline \textbf{Methods} &\textbf{GSM8K} & \textbf{AQuA} & \textbf{OpenbookQA} & \textbf{CR} \\\hline
            CoT & 97.93 & 88.98 & 93.21 & 78.92\\
            Self-rethinking &  \textbf{98.02} & \textbf{91.03} & \textbf{95.07} & \textbf{81.37} \\\hline
        \end{tabular}
        }
    \caption{GPT4' results on zero-shot Chain-of-Thought (CoT) and our methods, self-rethinking. }
    \label{tab:GPT4_res}
\end{table}

\subsection{Experiment Setup}
We conduct comparisons between self-rethinking and several other baselines on multiple benchmarks. 

\textbf{Baselines: } We select the following reasoning baselines to evaluate our framework, self-rethinking's performance. 
\begin{itemize}[before=\vspace{-0.5\baselineskip},itemsep=0mm]
    \item Standard prompting~\citep{brown2020language}: the basic reasoning promptings with prefixes as \textit{question} and \textit{answer}. 
    \item Chain-of-Thought~(CoT)~\citep{madaan2023self}: a technique that enhances large language models' ability to perform complex and multi-step reasoning by guiding them through a problem-solving process step by step, significantly improving their performance on tasks that require deeper cognitive processing.
    \item Self-refine~\citep{madaan2023self}:  an approach that enables LLMs to iteratively improve their initial outputs by providing feedback to themselves and refining their responses.
    \item Self-consistency~\citep{wang2022self}: a decoding strategy that enhances CoT prompting in LLMs by sampling multiple reasoning paths and selecting the most consistent answer.
\end{itemize}

\textbf{Benchmarks:} We consider the following existing math problems benchmarks designed with human rationale reference. 
\begin{itemize}[before=\vspace{-0.5\baselineskip},itemsep=0mm]
    \item GSM8K benchmark of math word problems~\citep{cobbe2021training}. 
    \item AQuA dataset of algebraic math problems~\citep{ling2017program}. 
    \item MathQA benchmark of multiple-choice math problems~\citep{amini2019mathqa}.
    \item Openbook benchmark modeled after open
    book exams for assessing human understanding of a subject~\citep{mihaylov2018can}. 
    \item LogiQA dataset sourced from expert-written questions for testing human logical reasoning~\citep{liu2020logiqa}. 
    \item Critical Reasoning in MARB benchmark of several graduate admission tests, highlighting the reasoning to assumptions, conclusions and paradoxes in arguments~\citep{tong2023eliminating}. 
\end{itemize}

\textbf{Models: } In order to evaluate self-rethinking's effects, we choose PaLM2~\citep{anil2023palm} and GPT4~\citep{openai2023gpt4} as the baseline model. 
PaLM2 is a dense left-to-right, decoder-only language model. 
It is pre-trained on a high-quality corpus of 780 billion tokens with filtered webpages, books, Wikipedia, news articles, source code, and social media conversations. 
GPT4 is a large-scale multi-modal state-of-the-art model that exhibits human-level performance on various tasks. 
We use PaLM2's \textsc{text-bison-001} and GPT4's \textsc{gpt-4} models provided in their APIs. 

For mistake tuning, we choose two different-sized Flan T5~\citep{chung2022scaling}, which are specifically designed for instruction tuning strategies. This model excels in understanding and generating human-like text, demonstrating remarkable performance across a wide range of natural language processing tasks. 

\textbf{Training Details: }All of the following experiments were designed with a common setting, employing a random seed of 42, learning rate=1e-4. Considering the vast number of data in AQuA, we only randomly select 10,000 of them to represent the differences in tuning on two different domains.

\subsection{Self-rethinking Results}
\label{sec: Self-rethinking res}
Table~\ref{tab: arithmetic reasoning res} presents PaLM2's evaluation results on chosen benchmarks. 
In this experiment, we set our method, self-rethinking's $k$ equal to 1 to trade between the accuracy and computing resources. 
In order to align the commuting budget with our methods, we set the times of inference in self-consistency to 3.
 Our approach involves an initial zero-shot CoT inference, then rethinking whether this rationale has made similar errors. This leads to the final answer if no errors are found. If inaccuracies are detected, it combines a demonstration and the previously suspected erroneous answer for a third inference to arrive at the final answer. Hence, the overall inference times in our methods are between 2 and 3 times per question, which is still lower than self-consistency here. 
 
With the considered computational settings, the self-rethinking method shows superior performance with significant improvements, especially in GSM8K, AQuA, MathQA, and LogiQA, clearly outperforming self-consistency under a similar computing cost. 
However, while our method surpasses CoT in performance on the MathQA dataset, it falls short of achieving self-refine results. 
It's important to note that this dataset is specifically tailored towards operation-based arithmetic problems rather than general questions, aiming to gauge the models' proficiency in tackling complex issues~\citep{amini2019mathqa}. 
This suggests that the nature of the MathQA dataset may inherently be more suitable for self-refine.
In contrast to our approach, which aims to amend responses by identifying and addressing typical errors.
Table~\ref{tab:GPT4_res} compares GPT4's performance of CoT and self-rethinking. The results demonstrate a notable improvement when using our self-rethinking method over CoT. These findings suggest that self-rethinking is a more effective approach for enhancing GPT-4's performance.

Table~\ref{tab:few_shot} presents the 8-shot examples of CoT and self-rethinking, using the PaLM2 model across four different tasks: GSM8K, AQuA, MathQA, and LogiQA.
A key part of the process involved collecting PaLM2’s incorrect rationales for these examples, which were then used as learning demonstrations to rethink.
The results show a clear advantage of the self-rethinking method over the standard 8-shot CoT approach. 
These results highlight the efficacy of the self-rethinking method in improving accuracy in few-shot learning scenarios for complex problem-solving tasks.

Notably, self-refine shares our basic motivations about self-refining or self-correcting their answers but without utilizing any mistake samples. 
The result shows that our self-rethinking outperformed self-refine by a considerable margin across most of the datasets. 
This indicates the importance of our proposal for utilizing previous mistake examples. 
While self-refine demonstrates improvements in three arithmetic reasoning datasets, it concurrently exhibits substantial performance drops in commonsense reasoning datasets. 
By contrast, our self-rethinking consistently outperforms the standard method in various domains. 
This further implies the introduction of previous mistakes can stabilize the refinement and rethinking process.

In conclusion, our self-rethinking method achieved remarkable accuracy improvements in most tests, particularly in scenarios that demand high logical rigor and offer the opportunity to learn from errors by identifying fixed logical patterns, especially in arithmetic reasoning tasks. 
It indicates self-rethinking effectiveness in tasks requiring strong logic and prone to minor errors.
Additionally, the self-rethinking method proves particularly beneficial in assisting LLMs in identifying and rectifying low-level mistakes or misunderstandings that are within the model's capabilities but have been previously overlooked. 
This capability indicates that self-rethinking can serve as a valuable tool in refining the accuracy and reliability of responses in LLMs, especially in complex problem-solving contexts.

\begin{table}[h]
    \small
    \centering
    \scalebox{0.8}{
    \begin{tabular}{c|c|ccc} \hline
         \textbf{Models} & \textbf{Methods} & \textbf{GSM8K} & \textbf{MathQA} & \textbf{AQuA} \\\hline
         Flan-T5-large & Standard finetuning &  14.28 & 42.79 & 13.10 \\ 
         \small{(780M)} & Mistake tuning & \textbf{18.36} & \textbf{48.95} & \textbf{18.07}  \\\hline
         Flan-T5-xl & Standard finetuning &  23.81 & 47.24 &17.81 \\ 
         \small{(3B)} & Mistake tuning & \textbf{24.29} & \textbf{52.22} & \textbf{20.99}  \\\hline
    \end{tabular} 
    }
    \caption{Accuracy of Standard finetuning models (with only correct rationales) vs. our methods, mistake tuning (combined correct and incorrect rationales). Mistake tuning shows consistent and superior performance compared with only fine-tuned correct rationales.}
    \label{tab: mistake tuning res}
\end{table}

\subsection{Mistake Tuning Results}
\label{sec: mt res}
Table~\ref{tab: mistake tuning res} showcases the performance of the Flan-T5 models in the context of mistake tuning, highlighting the impact of combining correct and incorrect rationales.
The data presented in Table~\ref{tab: mistake tuning res} reveals significant insights into the performance of Flan-T5 models under mistake tuning, which involves integrating both correct and incorrect rationales. This approach is evident across different model scales, whether it’s the smaller 780M version or the larger 3B variant. 
Notably, in the MathQA domain, Flan-T5-large(780M) tuned by our methods demonstrates superior performance compared to PaLM2, achieving an accuracy of 48.95\% versus 41.37\%.
This phenomenon suggests that LLMs can benefit from engaging with incorrect reasoning, thereby enhancing their problem-solving and reasoning capabilities.
It extends beyond merely bolstering the model's grasp of correct CoT, to also encompassing the ability to identify and learn from incorrect rationales.

Furthermore, the expense of obtaining ground truth or hand-crafted references is significantly higher compared to generating and collecting incorrect rationales. 
This cost disparity underscores the practical value of our approach, offering a more cost-effective solution without compromising the quality of training data for machine learning models.
All mentioned provides a direction for further work of reasoning, which involves not only enhancing the model's understanding and learning of correct CoT but also the ability to identify and learn from incorrect rationales.

\section{Further Studies}
\begin{figure}[h]
    \centering
    \scalebox{0.3}{\includegraphics{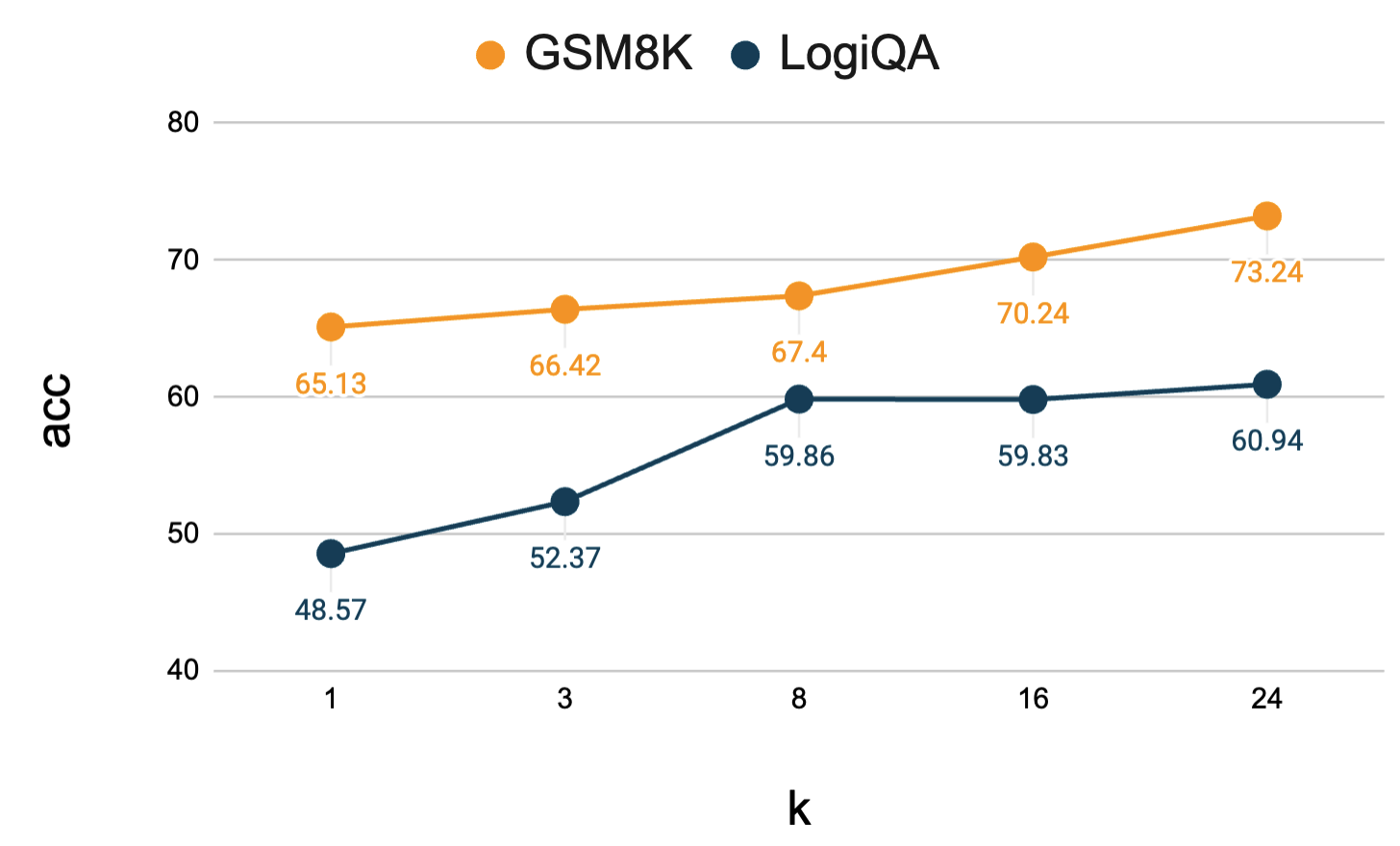}}
    \caption{Accuracy of different re-thinking iterations($k$). As the value of $k$ increases, the overall prediction accuracy improves.}
    \label{fig: iterations}
\end{figure}
\subsection{Hyperparameter Analysis of Rethinking Iteration Times}
In this section, we conduct experiments to assess the impact of different rethinking iterations, denoted as $k$, on the performance of our framework. 
We evaluate it on two mainstream benchmarks in the field of mathematics and commonsense reasoning, GSM8K and LogiQA. 
Figure~\ref{fig: iterations} represents the detailed trend under varying re-thinking times. 
Notably, as $k$ increases from 1 to 24, GSM8K represents a growth of 8.11\% and 12.37\% in LogiQA. 
It is evident as $k$ increases, both LLMs' arithmetic and commonsense reasoning accuracy exhibit an upward trend. 
This trend suggests a positive correlation between the number of rethinking iterations and the overall reasoning abilities. 
These observations indicate self-thinking's potential benefits with more inference time.   

\begin{table}[h]
\centering
\small
\begin{tabular}{@{}cccc|cc@{}}
\hline
CAT. & DEM. & COR. & INC. & \textbf{GSM8K} & \textbf{LogiQA} \\ \hline
 \checkmark    &      &      &      & 64.30 & 50.21  \\
 \checkmark    & \checkmark     &      &     & 62.70  & 48.57  \\
  \checkmark   & \checkmark     & \checkmark     &      & 65.70 & 51.01  \\
   \checkmark  &  \checkmark    &  \checkmark    &  \checkmark &   65.13  & 49.21 \\ \hline
\end{tabular}
\caption{Impact of Component Combinations. CAT. stands for the previous mistakes' type name, DEM. are the reasons for making such mistakes, and COR. and INC. mean corresponding correct and incorrect rationale examples. All components here are generated by LLM itself before reasoning. }
\label{tab: different_part}
\end{table}

\subsection{Ablation Study on Rethinking Process}
In this ablation study, we examined the impact of various component combinations in promptings to guide LLMs to self-rethinking . 
Table~\ref{tab: different_part} shows the performance of different components. 
The results indicate that the inclusion or exclusion of different components has varying effects on PaLM2's accuracy in domains of GSM8K and LogiQA. 
However, the overall performance across various components is relatively similar. 
It performs similarly well regardless of the specific combination of components, indicating good generalizability of the method.
This study suggests our method's flexibility and stability in future usage. 



\section{Unveiling LLM's Reasoning Errors}
\label{sec: ana}

\begin{figure}[t]
    \subfigure[Commonsense Reasnoing]{
        \includegraphics[width=0.51\linewidth]{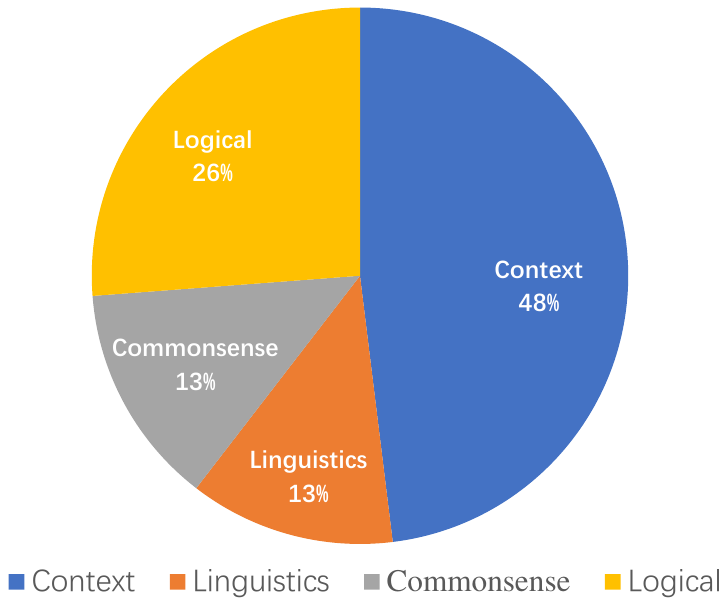}
    }
    \subfigure[Arithmetic Reasoning]{
        \includegraphics[width=0.39\linewidth]{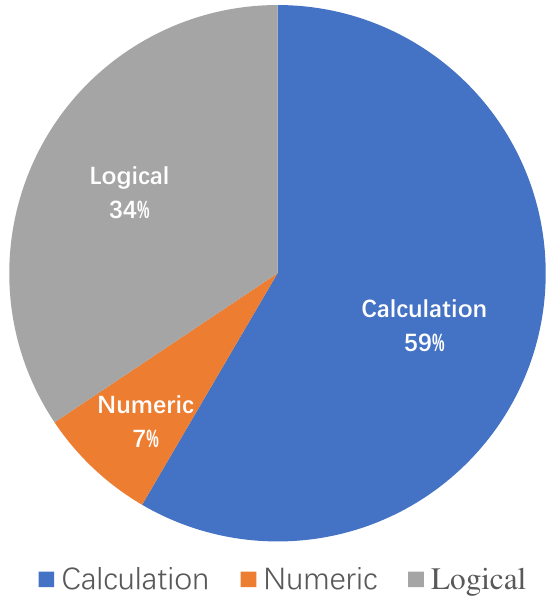}
    }
    \vspace{-3mm}
    \caption{PaLM2's error type distribution in the commonsense and arithmetic reasoning task.}
    \label{fig: error}
    \vspace{-5mm}
\end{figure}

In this section, we delve into the detailed types and underlying reasons that lead to mistakes in LLMs's inference process.
We sample mistake examples from GSM8K and LogiQA to conduct an in-depth analysis of both arithmetic and commonsense reasoning.
We put some examples in Appendix~\ref{sec: reasoning mistakes}.

For commonsense reasoning, we find errors like the misinterpretation of facts or concepts usually arise due to the model's limitations in understanding and applying context accurately.
This reveals current LLMs may still fall short of consistently recalling precise factual knowledge within a given context. 
Consequently, this underscores the imperative to advance toward the development of Retrieval-Augmented Generation(RAG) systems~\cite{guu2020retrieval,mallen2022not}, as they hold the promise of yielding more faithful and contextually aligned results.
Additionally, errors stemming from logical fallacies or incorrect inferences reveal LLMs' reliance on pattern recognition over logical reasoning, sometimes leading them to make logically inconsistent or unsupported connections by the given facts.



As shown in Figure~\ref{fig: error}, the most errors made by LLMs in arithmetic reasoning are about calculation.
This can be attributed to the different nature of LLMs compared to other tools like calculators.
To address this issue, \citet{chen2022program}'s suggestion using Program-of-Thought (PoT) is a promising approach to instruct LLMs to generate a segment of code to solve the given problem, resulting in more accurate calculation results.
Furthermore, it's important to note that logical error is also a type of error that LLMs always suffer from.
Compared with calculation errors and numeric errors, the causes of logical errors are more complicated and nuanced.
For instance, errors like misinterpreting given data or misapplying arithmetic operations reveal a lack of depth in understanding mathematical relationships.
This can result from the model's limitations in comprehending the nuances of mathematical concepts or its inability to correctly infer the needed function from the context of the question.
In the future, more fine-grained analysis and methods are needed to address such complex logical errors in arithmetic reasoning.

\section{Conclusions and Future Work}
In this work, we explore whether LLMs can learn from their mistakes. 
In order to investigate LLMs' abilities to differentiate and learn from mistakes, 
we introduce \textsc{CoTErrorSet}, a novel benchmark collecting both correct and incorrect CoT rationales across various domains and designed with demonstrations for making errors. 
We propose two possible solutions to expose the effects of mistakes from different perspectives: self-rethinking and mistake tuning. 
Both of them have achieved consistent and significant improvements, 
which demonstrates the potential benefits of learning from reasoning errors.
In the last, we conduct a comprehensive and detailed analysis of LLMs' common mistakes in both arithmetic and commonsense reasoning. 
The findings will provide a clear direction for future improvements.

For future work, we envision proposing corresponding algorithms or loss functions to learn implicit information from mistakes.
The primary intent of this work is to provide a new paradigm so there are still a lot of improvements that can be down following this work.
For example, incorporating contrastive learning to differentiate correct references and errors is intuitive to make more improvements. 
Also, some memorization and retrieval-augmented skills can help models benefit from mistakes similar to each question.

\section*{Limitations}
In addition to the noted challenge of fine-tuning commercial LLMs, we recognize several other specific limitations in our study that require attention. 
Primarily, our self-rethinking methodology may not be entirely suitable for tasks where a distinct, objective label is not readily available, such as in machine translation or dialogue generation. 
These areas pose a unique challenge as the correctness of outputs can often be subjective or context-dependent, making it difficult to apply our approach effectively.
Moreover, our utilization of the \textsc{CoTErrorSet} collection for mistake tuning necessitates a ground truth label for each sample, posing a potential impediment to the applicability of our method in low-resource scenarios.
In the future, we will continually improve our method and bring the concept of learning from mistakes to wider scenarios and applications. Thanks again for your thoughtful insights and informative comments.


\section*{Acknowledgements}
Our work is sponsored in part by NSF CAREER Award 2239440, NSF Proto-OKN Award 2333790, as well as generous gifts from Google, Adobe, and Teradata. Any opinions, findings, and conclusions or recommendations expressed herein are those of the authors and should not be interpreted as necessarily representing the views, either expressed or implied, of the U.S. Government. The U.S. Government is authorized to reproduce and distribute reprints for government purposes not withstanding any copyright annotation hereon.

\bibliography{custom}
\appendix
\onecolumn

\section{Algorithm for self-rethinking}
\begin{algorithm*}
    \begin{algorithmic}
        \STATE \( \textit{Mistakes} = \{ ...\} \)
        \STATE \( \textit{Correct \& Incorrect Examples} = \{ ...\} \)
        \STATE \textit{ErrorCounter} \( \leftarrow \) 0
        \STATE \textbf{Prompt:} Why you made the mistakes? 
        \STATE \textit{Mistakes} \( \leftarrow \) Error Type, Demonstrations, Examples. \\

        \STATE \textbf{Stage1 Prompt:} Let's think step by step. \\
        \STATE \textbf{Stage2 Prompt:} Do you make the same mistakes in \textit{Mistakes}? 
        
        \WHILE{ErrorCounter < k}
            \IF{Yes}
                \STATE go to Step2
                \STATE \textit{ErrorCounter} \( \leftarrow \) \textit{ErrorCounter} + 1
            \ELSIF{No}
                \STATE get the answer
                \STATE \textbf{break}
            \ENDIF
        \ENDWHILE

        \IF{\textit{ErrorCounter} == k}
            \STATE \textbf{Assume:} Problem or error detection exceeds the model's capabilities.
        \ENDIF

        \STATE \textbf{Prompt: }So the final answer is: 
    \end{algorithmic}
\caption{self-rethinking}
\end{algorithm*}


\section{Reasoning Mistake Examples}
\label{sec: reasoning mistakes}

\begin{table*}[t]
    \centering
    \small
    \label{tab:error_types}
    \noindent\rule{\textwidth}{1pt}
    \begin{itemize}[leftmargin=*,label={},itemsep=0pt]
    
    \item \textbf{Error name: Misinterpretation of Given Data} \\ 
    \textbf{Error type: Logical}
        \begin{itemize}
            \item \textit{Example:} Natalia sold clips to 48 of her friends in April, and then she sold half as many clips in May. How many clips did Natalia sell altogether in April and May?
            \item \textit{Correct Answer:} Natalia sold 48/2 = 24 clips in May. Natalia sold 48+24 = 72 clips altogether in April and May.
            \item \textit{Incorrect Rationale:} Natalia sold 48 * 2 = 96 clips in May. Natalia sold 48+96 = 144 clips altogether in April and May.
            \item \textit{Demonstration:} Mistaking multiplication for division led to a significant overestimate of the total clips sold.
        \end{itemize}

    \noindent\rule{\textwidth}{1pt}

    \textbf{Error type: Overlooking Details} \\
    \textbf{Error type: Logical}
    \begin{itemize}
        \item \textit{Example:} Mark has a garden with flowers. He planted plants of three different colors in it. Ten of them are yellow, and there are 80\% more of those in purple. There are only 25\% as many green flowers as there are yellow and purple flowers. How many flowers does Mark have in his garden?
        \item \textit{Correct Answer:} There are 80/100 * 10 = 8 more purple flowers than yellow flowers. So in Mark's garden, there are 10 + 8 = 18 purple flowers. Purple and yellow flowers sum up to 10 + 18 = 28 flowers. That means in Mark's garden there are 25/100 * 28 = 7 green flowers. So in total Mark has 28 + 7 = 35 plants in his garden.
        \item \textit{Incorrect Rationale:} There are 80/100 * 10 = 8 more purple flowers than yellow flowers. So in Mark's garden, there are 10 + 8 = 18 purple flowers. That means in Mark's garden there are 25/100 * 18 = 4.5 green flowers. So in total Mark has 10 + 18 + 4.5 = 32.5 plants in his garden.
        \item \textit{Demonstration:} Neglecting to consider both yellow and purple flowers in the green flower calculation led to a significant underestimation of the total number of flowers in Mark's garden.
    \end{itemize}
    \noindent\rule{\textwidth}{1pt}
    
    \textbf{Error name: Misapplication of Arithmetic Operation} \\
    \textbf{Error type: Calculation}
    \begin{itemize}
        \item \textit{Example:} Weng earns \$12 an hour for babysitting. Yesterday, she just did 50 minutes of babysitting. How much did she earn?
        \item \textit{Correct Answer:} Weng earns 12/60 = \$0.2 per minute. Working 50 minutes, she earned 0.2 x 50 = \$10.
        \item \textit{Incorrect Rationale:} Weng earns 12/60 = \$2 per minute. Working 50 minutes, she earned 2 x 50 = \$100.
        \item \textit{Demonstration:} Confusing the rate per hour with the rate per minute led to a substantial overestimation of earnings.
    \end{itemize}
    \noindent\rule{\textwidth}{1pt}


    \textbf{Error name: Numerical} \\
    \textbf{Error type: Numeric}
    \begin{itemize}
        \item \textit{Example:} The chicken crossed the road to get to the other side twice for the thrill of it. The first time, it had to dodge 23 speeding cars. The second time, a person tried to catch it and accidentally pulled out twice as many feathers as the number of cars the chicken had dodged. The chicken had 5263 feathers before its thrill-seeking road crossings. How many feathers did it have afterward?
        \item \textit{Correct Answer:} The chicken lost 23 * 2 = <<23*2=46>>46 feathers on its second road crossing.Thus, it had 5263 - 46 = <<5263-46=5217>>5217 feathers after crossing the road twice.
        \item \textit{Incorrect Rationale:} The chicken lost 23 * 2 = <<23*2=46>>46 feathers on its second road crossing. Thus, it had 5263 - 46 = <<5263-52=5211>>5211 feathers after crossing the road twice.
        \item \textit{Demonstration:} 1. The correct answer is 5217, while your answer is 5211. 2. Your answer is wrong because you subtracted 52 instead of 46. 3. The type name of the incorrect answer is numerical.
    \end{itemize}
    \noindent\rule{\textwidth}{1pt}
    
   \end{itemize}
\caption{Examples of Error Types in Arithmetic Reasoning. All contents are generated by PaLM2 itself.}
\end{table*}

\begin{table*}[h]
    \centering
    \label{tab:reasoning_error_types}
    \small
    \noindent\rule{\textwidth}{1pt}

    \begin{itemize}[leftmargin=*,label={},itemsep=0pt]



    \item \textbf{Error name: Logical Fallacy or Incorrect Inference}\\
    \textbf{Error type: Logical}
        \begin{itemize}
            \item \textit{Example:} "When standing miles away from Mount Rushmore"
            \item \textit{Correct Rationale:} Objects appear smaller when viewed from a greater distance.
            \item \textit{Incorrect Rationale:} "The mountains do not look smaller when standing miles away from Mount Rushmore. They look larger." (Logical fallacy)
            \item \textit{Demonstration:} 1. The correct rationale is that objects appear smaller when viewed from a greater distance, whereas the incorrect rationale states the opposite. 2. This is a logical fallacy as it contradicts a basic principle of perception. 3. The type name of the incorrect rationale is logical.
        \end{itemize}
        \noindent\rule{\textwidth}{1pt}
        \item \textbf{Error name: Incorrect Assumptions or Generalizations}\\
    \textbf{Error type: Logical}
        \begin{itemize}
            \item \textit{Example:} "Poison causes harm to which of the following?"
            \item \textit{Correct Rationale:} Poison affects living organisms.
            \item \textit{Incorrect Rationale:} "Robots do not get hurt by poison." (Incorrect generalization about the effects of poison)
            \item \textit{Demonstration:} 1. The correct rationale is that poison affects living organisms, but the incorrect rationale generalizes that robots are immune to poison. 2. This is an incorrect generalization because robots, being non-living entities, are not subject to biological effects. 3. The type name of the incorrect rationale is logical.
        \end{itemize}

    \noindent\rule{\textwidth}{1pt}

    \item \textbf{Error name: Misunderstanding Literal vs. Metaphorical Language}\\
    \textbf{Error type: Linguistics}
        \begin{itemize}
            \item \textit{Example:} "When food is reduced in the stomach"
            \item \textit{Correct Rationale:} Digestion involves the breakdown of food by stomach acid.
            \item \textit{Incorrect Rationale:} "Choice D is incorrect because it is not a fact." (Misunderstanding metaphorical language)
            \item \textit{Demonstration:} 1. The correct rationale is about the literal process of digestion, whereas the incorrect rationale misinterprets the metaphorical language. 2. This demonstrates a misunderstanding of metaphorical language. 3. The type name of the incorrect rationale is linguistics.
        \end{itemize}


    \noindent\rule{\textwidth}{1pt}

    \item \textbf{Error name: Factual Inaccuracy}\\ 
    \textbf{Error type: Commonsense}
        \begin{itemize}
            \item \textit{Example:} "You can make a telescope with a"
            \item \textit{Correct Rationale:} A telescope requires specific optical elements to function.
            \item \textit{Incorrect Rationale:} "A telescope needs a lens and a magnifying glass is a lens, so glass is a good choice." (Factually inaccurate about how telescopes are made)
            \item \textit{Demonstration:} 1. The correct rationale is that a telescope requires specific optical elements, whereas the incorrect rationale assumes any lens, like a magnifying glass, can make a telescope. 2. This shows a factual inaccuracy in understanding how telescopes are constructed. 3. The type name of the incorrect rationale is commonsense.
        \end{itemize}
    
    \noindent\rule{\textwidth}{1pt}

\item \textbf{Error type: Misunderstanding Context or Relevance} \\
    \textbf{Error type: Context}
    
        \begin{itemize}
            \item \textit{Example:} "an inherited characteristic found on all mammals is"
            \item \textit{Correct Rationale:} Inherited characteristics in mammals include features like fur.
            \item \textit{Incorrect Rationale:} "Shoes are not found on all mammals" (Misunderstanding the context of biological characteristics)
            \item \textit{Demonstration: } 1. The correct rationale focuses on relevant inherited physical traits like fur. 2. This error illustrates a clear lack of understanding of the context. 3. The type name of the incorrect rationale should be context.   \\

        \end{itemize}

    \noindent\rule{\textwidth}{1pt}
    \end{itemize}
\caption{Examples of Error Types in Commonsense Reasoning. All contents are generated by PaLM2 itself.}
\end{table*}

\clearpage

\section{More Details about LLM-based Clustering Approach}
\label{Definition for Error Categories}

\begin{table*}[h]
\centering
\begin{tabular}{ll}
\hline
Input  & \begin{tabular}[c]{@{}l@{}}Please generate several keywords to cover all the following error types, and \\ assign each keyword to an error type category. Output in the following format:\\ {[}Specific Error Category1{]}: {[}keyword1{]}, {[}keyword2{]}\\ {[}Specific Error Category2{]}: {[}keyword3{]}, {[}keyword4{]}\\ Keywords: \{keywords\}\end{tabular} \\
\hline
Output & \begin{tabular}[c]{@{}l@{}}Mathematical: \{keywords cluster1\}\\ Numerical: \{keywords cluster2\}\\ Arithmetic: \{keywords cluster3\}\\ Calculation: \{keywords cluster4\}\end{tabular}                                                                 \\
\hline
\end{tabular}
\caption{Detailed input and output of our LLM-based clustering method.}
\end{table*}

\begin{table*}[h]
\centering
\begin{tabular}{lp{4.5in}p{4.5in}}
\hline
Error Type  & Definition \\ \hline
Calculation & Mistakes or inaccuracies that occur during the process of performing mathematical calculations. These errors can arise from various sources and can occur at any stage of a mathematical problem-solving process.                                                                                                                                                      \\\hline
Numeric     & Numeric errors in the context of mathematical reasoning refer to inaccuracies that arise from the representation and manipulation of numerical values. These errors can occur at various stages of mathematical computations and can result from limitations in the precision of the representation of real numbers or mistakes in handling numerical data.            \\\hline
Logical     & Logical errors involve mistakes in the overall reasoning or strategy used to solve a mathematical problem. This type of error may not be immediately apparent during the calculation process but can lead to incorrect final results. It could include using an incorrect formula or assumptions, misunderstanding the problem statement, or applying the wrong concept. \\\hline
Linguistics & Errors in linguistics involve inaccuracies or mistakes in the use of language. These can include grammatical errors, misuse of vocabulary, incorrect syntax, or problems in semantics. Linguistic errors may arise from a lack of understanding of the rules of a language, misinterpretation of meaning, or the inability to effectively convey a message in a given language. Such errors can affect the clarity, coherence, and overall effectiveness of communication. \\\hline
Commonsense & Commonsense errors refer to mistakes or inaccuracies that occur in the application of general world knowledge or everyday reasoning. These errors can arise from misconceptions, flawed logic, or misunderstandings of basic principles that are widely accepted as common knowledge. Commonsense errors often lead to conclusions or decisions that, upon closer examination, are illogical or inconsistent with general understanding of the world. \\\hline 
Context & Errors of misunderstanding context or relevance occur when there is a failure to correctly interpret or apply the relevant information in a given scenario. This type of error typically involves overlooking key aspects of a context, making inappropriate generalizations, or failing to distinguish between literal and metaphorical language. These errors can significantly alter the intended meaning or relevance of a response in reasoning tasks. \\ \hline

\end{tabular}
\caption{PaLM2's Understanding and Definitions for Error Types. All contents are generated by itself after providing its mistakes and corresponding golden-standard references.}
\end{table*}

\end{document}